\documentclass[11pt]{article}

\usepackage[final]{acl}

\usepackage{times}
\usepackage{latexsym}

\usepackage[T1]{fontenc}
\usepackage[utf8]{inputenc}

\usepackage{microtype}

\usepackage{inconsolata}

\usepackage{graphicx}
\usepackage{enumitem}
\usepackage{hyperref}
\usepackage{natbib}
\usepackage{svg}
\usepackage{multirow}
\usepackage{listings}
\usepackage[normalem]{ulem}
\useunder{\uline}{\ul}{}

\title{Kakugo: Distillation of Low-Resource Languages into Small Language Models}

\author{Peter Devine, \phantom{aa} Mardhiyah Sanni, \phantom{aa} Farid Adilazuarda, \\ \textbf{Julieta Gil Loizaga, \phantom{aa}Barry Haddow} \\
  School of Informatics, University of Edinburgh \\
  Edinburgh, UK \\
  \texttt{\{pdevine2, msanni, farid.adilazuarda, julieta.loizaga, bhaddow\}@ed.ac.uk}}

\begin{document}
\maketitle
\begin{abstract}

We present Kakugo, a novel and cost-effective pipeline designed to train general-purpose Small Language Models (SLMs) for low-resource languages using only the language name as input. By using a large teacher model to generate synthetic prompts and translate instruction datasets, we produced training data and SLMs for 54 low-resource languages. Evaluations across a diverse set of general natural language processing tasks, including translation, classification, and question answering, demonstrate that our pipeline consistently improves performance over base models. With a total generation and training cost of under \$50 per language, Kakugo offers an accessible method for communities to develop language-specific AI.

\end{abstract}

\section{Introduction}

Small Language Models (SLMs) offer a cheaper, faster, and more data-sovereign alternative to Large Language Models (LLMs)~\cite{schick2021s}, achieving remarkable performance across diverse NLP tasks~\cite{abdin2024phi,chae2025large,zheng2025hunyuan,pham2025slimlm,li2025leveraging}. However, this efficiency often comes at the cost of multilingual capability. While popular SLMs support many languages~\cite{gemma_2025,qwen3technicalreport,bakouch2025smollm3,swissai2025apertus}, they typically underperform in low-resource settings compared to their larger counterparts~\cite{romanou2024include}. Although the community has attempted to fill this void with language-specific SLMs~\cite{awagptv1_2025,ogueji-etal-2021-small,2504.05747,tonja2024inkubalm,khanuja2021muril}, these efforts often require expensive specialist teams and intimate knowledge of data sources.

To democratize access to language-specific models, we present \textbf{Kakugo}, a general-purpose pipeline that trains a monolingual SLM using \textit{only} the language name as input.

\begin{figure*}[t]
    \centering
    \includegraphics[width=\textwidth]{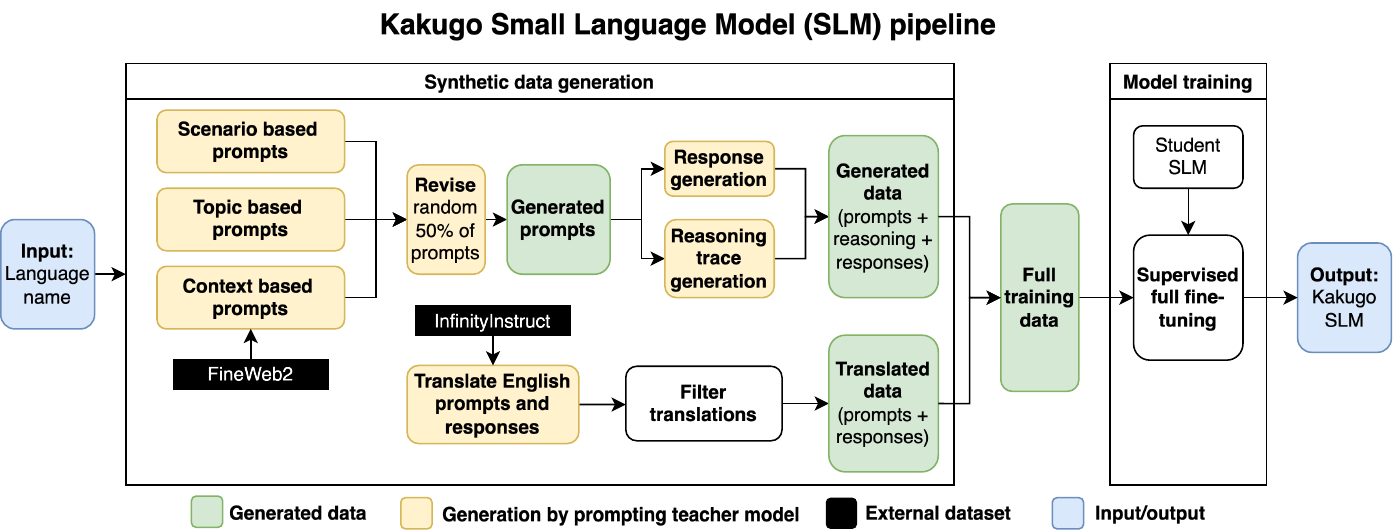}
    \caption{The Kakugo pipeline. Given a language name (e.g., ``Javanese''), a teacher model generates synthetic conversational and instruction data which are used to train an SLM in that language.}
    \label{fig:kakugo_pipeline}
\end{figure*}

Kakugo uses model distillation~\cite{hinton2015distilling,shirgaonkar2024knowledge,wang2021towards} to automate the entire development process. We use a large teacher model, specifically GPT-OSS 120B~\cite{openai2025gptoss120bgptoss20bmodel}, to generate diverse synthetic prompts, reasoning traces, and responses, while also translating high-quality instruction datasets into the target language. This data is then used to fine-tune a student model (IBM's Granite 4 Micro~\cite{granite2025}).

We evaluate Kakugo on 54 low-resource languages across translation~\citep[FLORES][]{nllb-24}, classification~\citep[SIB-200][]{adelani2024sib}, and question answering~\citep[GlobalMMLU][]{singh2025global}~\citep[Belebele][]{bandarkar2024belebele}. Our experiments demonstrate that our pipeline consistently improves performance over base models across a wide range of tasks.

We make three key contributions:

Firstly, we propose a novel, fully automated pipeline for generating generalist synthetic data, demonstrating that combining generated reasoning traces with translated data is more effective for SLM training than any individual method. 

Secondly, we release open-source training datasets and monolingual SLMs for 54 languages. Notably, for many of these languages (e.g., Bashkir, Maori, Mongolian), these represent the first generalist monolingual conversational SLMs ever developed. 

Thirdly, we release the Kakugo pipeline itself, enabling the creation of language-specific SLMs for less than \$50 per language. This allows low-resource communities to affordably develop AI tools that support their native languages.

The code for reproducing our pipeline, as well as links to the models and datasets produced in this work can be found at \href{https://github.com/Peter-Devine/kakugo}{https://github.com/Peter-Devine/kakugo}.

\section{Related work}

LLMs are effective data generators for downstream tasks like intent classification, recommendation, and inference \cite{dai2025auggpt, sahu2022data, liu2024once}. Distilling LLM outputs into SLMs significantly reduces inference costs while maintaining performance \cite{shirgaonkar2024knowledge}.

Synthetic data generation approaches range from simple frameworks such as Self-Instruct~\cite{wang2023self} and Alpaca~\cite{taori2023alpaca}, which use input-output mimicry, to more complex approaches such as Explanation Tuning in Orca~\cite{mukherjee2023orca} and Distilling Step-by-Step~\cite{hsieh2023distilling} which demonstrate that supervising student models on reasoning traces improves performance and data efficiency. These forms of model distillation have been shown to be very useful for reducing inference time and cost while maintaining performance across many tasks~\cite{shirgaonkar2024knowledge}. 

Notably, Nemotron-4 was trained on 98\% synthetic data and achieved improved evaluation scores in a broad array of tasks~\cite{adler2024nemotron}, though this model was not specifically optimized for low-resource languages.

In the multilingual domain, models such as Goldfish~\cite{chang2024goldfish}, EMMA-500~\cite{ji2024emma}, and MaLA-500~\cite{lin2024mala} have shown that pre-training on low resource language web text can improve down-stream task performance. Moreover, instruction tuning projects like BLOOMZ~\cite{muennighoff2023crosslingual} and Aya~\cite{ustun2024aya} have highlighted the efficacy of cross-lingual generalization through multitask fine-tuning. While these efforts rely heavily on human-curated or translated datasets, emerging synthetic methods for low-resource languages have explored ``reverse instruction'' generation, as seen in Humpback~\cite{li2023self} and MURI~\cite{koksal2024muri}, which create instructions from native web text to mitigate translation artifacts.

This approach of generating training data using an LLM has proved effective for training models in languages besides English, including low resource languages, for various domains such as legal NLP~\cite{ghosh2023dale}, translation~\cite{yong2024lexc}, text classification~\cite{anikina2025rigorous}, fact checking~\cite{chung2025beyond}, question answering~\cite{namboori2023gemquad}, and dialogue summarization~\cite{suresh2025diasynth}. Moreover, pre-training on synthetically generated data has shown to improve down-stream evaluation results~\cite{wang2025multilingual}.

However, none of these approaches result in a generalist low-resource language model which can be used by a wide variety of AI developers. Our work seeks to distill the generalist abilities of an LLM into an SLM for a given language without any strict task-based restrictions.

\section{Kakugo model generation pipeline}

The goal of this research was to build a universal pipeline where developers can produce target-language SLMs simply by specifying the language name.

Our pipeline works in two main steps: training data creation and model training. This section details how we both create our data and then use it to train our model.

\subsection{Data creation}
\label{generatingdata}

To create training data, we use a teacher model that is able to generate output in the target language. 
We select GPT-OSS 120B~\cite{openai2025gptoss120bgptoss20bmodel} as our teacher. Preliminary qualitative analysis confirmed its proficiency in generating coherent text across our target languages, validating its suitability for distillation without extensive few-shot tuning.

With this model, we generate data in two main ways: (1) generating prompts and then responses based on those prompts, and (2) translating existing English conversation data into the target language.

\subsubsection{Generating synthetic prompts}

We generate our prompts using three different methods: topic-based, scenario-based, and context-based prompt generation. We describe these methods in detail below.

\paragraph{Topic-based prompt generation}

In order to generate training data on a wide range of topics, both general and language specific, we take inspiration from the Nemotron 4~\cite{adler2024nemotron} prompt generation pipeline. We start with a set of seed topics then prompt the teacher model to automatically generate 20 macro-topics related to each seed and then 10 topics related to each macro-topic.
In order to include both general and language or culture specific data in our training, we use two sets of seeds: 8 language-agnostic seeds (e.g. ``practical skills'', ``sciences'') and 8 language-specific seeds (e.g. ``\{$language\_name$\} speaking places'', ``\{$language\_name$\} society''). The teacher LLM is then allowed to interpret what is meant by these seeds within the confines of the topic-generating system message (listed in Appendix~\ref{sec:seeds_and_sys_prompts}).
We then combine all seeds, macro-topics, and topics as a single set of topics which we use to prompt the teacher model to generate 3 conversational prompts in the target language for each topic. 
    
\paragraph{Scenario-based prompt generation}

To generate practical training data for realistic situations where an SLM will be used, we generate prompts based on specific scenarios.
We first prompt the teacher model to generate 30 broad, general scenarios in which one could imagine someone using an AI assistant. We generated both language-specific and language-agnostic scenarios by either informing or not informing the teacher model that the user is a speaker of the target language. 
We then generate 30 detailed scenarios related to each broad scenario. We define our final set of scenario seeds as the union of the 30 broad scenarios and the generated detailed scenarios. We iterate through this combined pool, prompting the teacher model to generate 5 target-language prompts for each individual scenario.

\paragraph{Context-based prompt generation}

We anticipate that low-resource language SLMs will be used for many text-based tasks such as translation or summarization, so we generate prompts that relate to a piece of text in some way.
We first randomly sample up to 10,000 texts in the target language from the popular FineWeb2 corpus~\cite{penedo2025fineweb2pipelinescale}. Taking the first 1,000 tokens of each text, we prompt the teacher model, through the system message, to generate up to 3 prompts that ask an AI assistant to do one of the following randomly selected tasks: to translate, summarize, improve, classify, or answer a question about the text. These tasks are uniformly distributed except the question answering task, which we over-weight by a factor of 4 due to its potential usefulness in  deploying our final model for RAG pipelines~\cite{lewis2020retrieval}.
To create our final context-based prompts, we prepended the text that was used to generate each prompt to the prompt itself.\\

Similarly to Nemotron~\cite{adler2024nemotron}, our initial tests showed that the generated prompts using all three methods were often simplistic or short. Therefore, we randomly sample half of each set of prompts and prompt the teacher model to make them longer or more complex. 
% For the half of prompts where a complex prompt is generated from a simple prompt, the complex prompt is used instead of the original simple one, which results in a dataset of half original prompts and half revised prompts. We chose half as a suitable proportion between original and revised prompts because it gives equal exposure to the student model of both short, simple prompts and long, complex prompts.

\subsubsection{Generating responses from prompts}

We input all of our synthetic prompts into our teacher model with a standard system message including instructions  that the model must reply in the target language unless otherwise instructed. 

Since we use a reasoning-based teacher model (GPT-OSS 120B) which automatically produces a `chain-of-thought' reasoning trace before its final answer, we capture these naturally occurring reasoning traces alongside the final response without requiring specific engineering of the system prompt to elicit them.

% Since we use a reasoning-based model as the teacher model in our experiments, this step generates both a reasoning trace and response for each prompt.

More information on the exact seeds and system prompts we used to create our generated prompts and responses can be found in Appendix~\ref{sec:seeds_and_sys_prompts}. 

\subsubsection{Translating data}

To increase the diversity of our synthetic data and to include more general instruction following data, we also generate data by translating an existing high quality English instruction dataset. We select the ``7M\_core'' (1.48M row) subset of the InfinityInstruct~\cite{li2025infinity} dataset which is a curated and filtered set of conversational English and Chinese prompt and response data based on many high-quality instruction datasets. We filter out non-English examples and select the first 15,000 rows of the dataset.

We translate this data as Python lists of conversations using our teacher model with the system message given in Appendix~\ref{sec:translation_sys_msg}, then parse these translated string responses back to Python lists, removing any instances that were incorrectly formatted.

To ensure data quality, we remove conversations where the translated token count is less than 75\% or exceeds 25 times the original length. These asymmetric filtering thresholds prevent omitting correctly translated data by accommodating the high ``token fertility'' and inefficient tokenization often observed in low-resource languages or rare scripts~\cite{rust2021good, ahia2023all}. While translation models are prone to repetition~\cite{mi2016coverage}, our requirement for correctly formatted Python lists in the translated output effectively excludes such instances, as repetitive strings are unlikely to maintain the syntax necessary to pass our formatting checks, making our 25 times length threshold effective at only filtering out truly repetitive conversations.

We use this filtered data as our translated training data.

\subsection{Model training}

For our final dataset, we combine our generated and translated data into a single shuffled dataset to train on. 

We structure our training data to add controllable reasoning capabilities to the student model. For the generated subset (which contains reasoning traces), we prepend a specific system instruction enabling ``thinking mode''. For the translated subset (which lacks reasoning traces), we use a standard system instruction. This allows the student model to learn a conditional association: it generates <think> </think> traces and reasoning steps only when the 'thinking mode' system prompt is present, effectively distilling the teacher's reasoning ability into a toggleable feature, similar to Qwen 3's ``thinking mode'' and ``non-thinking mode''~\cite{qwen3technicalreport}.

We train our student model on this data using Llama Factory~\cite{zheng2024llamafactory} by full fine-tuning for one epoch with the intention that we strengthen the generalist abilities of our model without overfitting to our training data. Since full finetuning frequently outperforms low rank training approaches (e.g. LoRA) when the trained model is applied to multiple tasks~\cite{shuttleworth2024lora}, we choose to do full finetuning as our training method in order to create a more generalist model.

In our experiments, we chose to train Granite 4 Micro~\cite{granite2025} as our student model due to its open-source license and its superior performance on evaluation benchmarks compared to similarly sized models.

The full set of training hyperparameters can be found in Appendix~\ref{sec:training_hyperparams}. 

\section{Experiments}

In order to compare the effects of the different data generation techniques in our pipeline, we generated data and used a variety of data subsets to train models on a fixed token training budget to evaluate them on our low-resource language evaluation benchmarks.

\subsection{Language selection}

We targeted a wide variety of languages balancing linguistic diversity with evaluation benchmark coverage. We selected the following 14 languages for in-depth evaluation across our data subsets: Amharic, Bengali, Galician, Guarani, Javanese, Kyrgyz, Lao, Maori, Mongolian, Scottish Gaelic, Sinhala, Swahili, Telugu, and Yoruba.

We also trained models in a further 40 languages, but we performed less in-depth analyses of these models, only training a single ``full'' model for each. These languages were selected by finding languages which are both included in FLORES 200 and have between 50,000 and 1,000,000 documents in FineWeb2. This was done to target languages which are low resource, but not so low resource that our teacher model could not reliably generate training data in that language. 

These 40 languages were: Aranese, Assamese, Asturian, Bashkir, Cebuano, Central Kurdish, Chuvash, Eastern Yiddish, Egyptian Arabic, Faroese, Haitian Creole, Hausa, Igbo, Irish, Kinyarwanda, Lhasa Tibetan, Luxembourgish, Maltese, Mizo, Najdi Arabic, Northern Kurdish, Nyanja, Papiamento, Plateau Malagasy, Rundi, Samoan, Shona, Sindhi (Arabic script), South Azerbaijani, Southern Pashto, Southern Sotho, Sundanese, Tajik, Tatar, Tigrinya, Turkmen, Uyghur, Welsh, Xhosa, Zulu.

This resulted in the creation of datasets and models for 54 low-resource languages.
\begin{table}[]
\begin{tabular}{|l|l|l|}
\hline
\textbf{} & \textbf{Num langs} & \textbf{Task} \\ \hline
\textbf{Belebele} & 34 & Closed QA \\ \hline
\textbf{GlobalMMLU} & 10 & Open QA \\ \hline
\textbf{SIB200} & 53 & Classification \\ \hline
\textbf{FLORES xx-en} & 54 & Translation \\ \hline
\textbf{FLORES en-xx} & 54 & Translation \\ \hline
\end{tabular}
\caption{Evaluation tasks and how many languages are covered by each task in our evaluation.}
\label{tab:eval_tasks}
\end{table}

% \usepackage{graphicx}
\begin{table*}[]
\centering
\resizebox{\textwidth}{!}{%
\begin{tabular}{|l|l|l|l|l|l|}
\hline
 & \textbf{Belebele} & \textbf{Global MMLU} & \textbf{SIB200} & \textbf{FLORES  xx-en} & \textbf{FLORES en-xx} \\ \hline
\textbf{Base model} & 41.8 & 33.7 & 66.5 & 37.6 & 20.0 \\ \hline
\textbf{Topic only*} & 39.8 & 34.4 & 65.6 & 29.9 & 18.5 \\ \hline
\textbf{Scenario only*} & 39.4 & 34.1 & 64.7 & 31.7 & 18.5 \\ \hline
\textbf{Context only*} & 43.1 & 33.5 & 67.1 & 36.3 & 18.1 \\ \hline
\textbf{Gen (Scen+Cont+Topic)*} & 42.8 & 34.4 & 66.7 & 34.2 & 18.8 \\ \hline
\textbf{Translated only*} & 41.6 & 34.0 & 62.2 & 32.8 & 18.3 \\ \hline
\textbf{Gen + Tran*} & 42.4 & 34.1 & 65.5 & 35.3 & 18.4 \\ \hline
\textbf{Gen (w. Reasoning) + Tran (token limited)*} & 41.7 & 33.5 & 65.6 & 38.3 & 18.2 \\ \hline
\textbf{Gen (w. Reasoning) + Tran (example limited)} & 42.3 & 33.3 & 66.2 & 38.9 & 19.0 \\ \hline
\textbf{Gen (w. Reasoning) + Tran (full)} & \textbf{48.8} & \textbf{35.3} & \textbf{74.9} & \textbf{44.9} & \textbf{30.7} \\ \hline
\end{tabular}%
}
\caption{Average score across our 14 ``in-depth'' languages for each dataset. Reported scores are percentage accuracy for all tasks except FLORES tasks, which are CHRF++ scores. Models marked * have been trained on the same number of tokens for each language.}
\label{tab:full-task-model-comparison}
\end{table*}
% \usepackage{multirow}
% \usepackage{graphicx}
% \usepackage[normalem]{ulem}
\begin{table*}[]
\centering
\resizebox{\textwidth}{!}{%
\begin{tabular}{lllrrrrrrrrrr}
\cline{4-13}
 &  & \multicolumn{1}{l|}{} & \multicolumn{1}{l|}{\textbf{Base}} & \multicolumn{1}{l|}{\textbf{Topic}} & \multicolumn{1}{l|}{\textbf{Scen}} & \multicolumn{1}{l|}{\textbf{Cont}} & \multicolumn{1}{l|}{\textbf{Gen}} & \multicolumn{1}{l|}{\textbf{Tran}} & \multicolumn{1}{l|}{\textbf{Gen + Tran}} & \multicolumn{1}{l|}{\textbf{\begin{tabular}[c]{@{}l@{}}GenReas + \\ Tran (tok-lim)\end{tabular}}} & \multicolumn{1}{l|}{\textbf{\begin{tabular}[c]{@{}l@{}}GenReas + \\ Tran (ex-lim)\end{tabular}}} & \multicolumn{1}{l|}{\textbf{\begin{tabular}[c]{@{}l@{}}GenReas + \\ Tran (full)\end{tabular}}} \\ \cline{4-13} 
 &  &  & \multicolumn{1}{l}{} & \multicolumn{1}{l}{} & \multicolumn{1}{l}{} & \multicolumn{1}{l}{} & \multicolumn{1}{l}{} & \multicolumn{1}{l}{} & \multicolumn{1}{l}{} & \multicolumn{1}{l}{} & \multicolumn{1}{l}{} & \multicolumn{1}{l}{} \\ \cline{1-1} \cline{3-13} 
\multicolumn{1}{|l|}{\multirow{14}{*}{\textbf{SIB200}}} & \multicolumn{1}{l|}{} & \multicolumn{1}{l|}{\textbf{Amharic}} & \multicolumn{1}{r|}{26.4} & \multicolumn{1}{r|}{{\ul 27.9}} & \multicolumn{1}{r|}{26.9} & \multicolumn{1}{r|}{24.9} & \multicolumn{1}{r|}{26.9} & \multicolumn{1}{r|}{26.4} & \multicolumn{1}{r|}{27.4} & \multicolumn{1}{r|}{26.9} & \multicolumn{1}{r|}{29.4} & \multicolumn{1}{r|}{\textbf{51.2}} \\ \cline{3-13} 
\multicolumn{1}{|l|}{} & \multicolumn{1}{l|}{} & \multicolumn{1}{l|}{\textbf{Bengali}} & \multicolumn{1}{r|}{85.6} & \multicolumn{1}{r|}{83.1} & \multicolumn{1}{r|}{85.1} & \multicolumn{1}{r|}{79.6} & \multicolumn{1}{r|}{88.6} & \multicolumn{1}{r|}{83.6} & \multicolumn{1}{r|}{{\ul \textbf{86.6}}} & \multicolumn{1}{r|}{84.6} & \multicolumn{1}{r|}{83.1} & \multicolumn{1}{r|}{85.6} \\ \cline{3-13} 
\multicolumn{1}{|l|}{} & \multicolumn{1}{l|}{} & \multicolumn{1}{l|}{\textbf{Galician}} & \multicolumn{1}{r|}{85.6} & \multicolumn{1}{r|}{{\ul 85.6}} & \multicolumn{1}{r|}{84.1} & \multicolumn{1}{r|}{85.1} & \multicolumn{1}{r|}{83.1} & \multicolumn{1}{r|}{71.6} & \multicolumn{1}{r|}{77.1} & \multicolumn{1}{r|}{81.1} & \multicolumn{1}{r|}{76.6} & \multicolumn{1}{r|}{\textbf{87.1}} \\ \cline{3-13} 
\multicolumn{1}{|l|}{} & \multicolumn{1}{l|}{} & \multicolumn{1}{l|}{\textbf{Guarani}} & \multicolumn{1}{r|}{69.2} & \multicolumn{1}{r|}{{\ul 70.6}} & \multicolumn{1}{r|}{66.7} & \multicolumn{1}{r|}{68.2} & \multicolumn{1}{r|}{69.2} & \multicolumn{1}{r|}{62.7} & \multicolumn{1}{r|}{69.2} & \multicolumn{1}{r|}{68.7} & \multicolumn{1}{r|}{67.7} & \multicolumn{1}{r|}{\textbf{77.1}} \\ \cline{3-13} 
\multicolumn{1}{|l|}{} & \multicolumn{1}{l|}{} & \multicolumn{1}{l|}{\textbf{Javanese}} & \multicolumn{1}{r|}{72.6} & \multicolumn{1}{r|}{77.1} & \multicolumn{1}{r|}{80.1} & \multicolumn{1}{r|}{{\ul 81.1}} & \multicolumn{1}{r|}{77.6} & \multicolumn{1}{r|}{75.1} & \multicolumn{1}{r|}{74.6} & \multicolumn{1}{r|}{79.1} & \multicolumn{1}{r|}{77.1} & \multicolumn{1}{r|}{\textbf{83.1}} \\ \cline{3-13} 
\multicolumn{1}{|l|}{} & \multicolumn{1}{l|}{} & \multicolumn{1}{l|}{\textbf{Kyrgyz}} & \multicolumn{1}{r|}{75.1} & \multicolumn{1}{r|}{76.1} & \multicolumn{1}{r|}{72.6} & \multicolumn{1}{r|}{78.1} & \multicolumn{1}{r|}{76.6} & \multicolumn{1}{r|}{{\ul 79.1}} & \multicolumn{1}{r|}{76.1} & \multicolumn{1}{r|}{76.1} & \multicolumn{1}{r|}{71.1} & \multicolumn{1}{r|}{\textbf{81.1}} \\ \cline{3-13} 
\multicolumn{1}{|l|}{} & \multicolumn{1}{l|}{} & \multicolumn{1}{l|}{\textbf{Lao}} & \multicolumn{1}{r|}{63.2} & \multicolumn{1}{r|}{47.3} & \multicolumn{1}{r|}{{\ul 52.2}} & \multicolumn{1}{r|}{49.8} & \multicolumn{1}{r|}{47.3} & \multicolumn{1}{r|}{{\ul 52.2}} & \multicolumn{1}{r|}{51.7} & \multicolumn{1}{r|}{51.7} & \multicolumn{1}{r|}{52.7} & \multicolumn{1}{r|}{\textbf{67.7}} \\ \cline{3-13} 
\multicolumn{1}{|l|}{} & \multicolumn{1}{l|}{} & \multicolumn{1}{l|}{\textbf{Maori}} & \multicolumn{1}{r|}{49.3} & \multicolumn{1}{r|}{50.2} & \multicolumn{1}{r|}{48.8} & \multicolumn{1}{r|}{{\ul 62.2}} & \multicolumn{1}{r|}{53.7} & \multicolumn{1}{r|}{46.3} & \multicolumn{1}{r|}{53.2} & \multicolumn{1}{r|}{57.2} & \multicolumn{1}{r|}{57.2} & \multicolumn{1}{r|}{\textbf{68.2}} \\ \cline{3-13} 
\multicolumn{1}{|l|}{} & \multicolumn{1}{l|}{} & \multicolumn{1}{l|}{\textbf{Mongolian}} & \multicolumn{1}{r|}{63.2} & \multicolumn{1}{r|}{66.2} & \multicolumn{1}{r|}{67.2} & \multicolumn{1}{r|}{69.7} & \multicolumn{1}{r|}{70.6} & \multicolumn{1}{r|}{65.7} & \multicolumn{1}{r|}{{\ul 72.6}} & \multicolumn{1}{r|}{{\ul 72.6}} & \multicolumn{1}{r|}{70.1} & \multicolumn{1}{r|}{\textbf{81.1}} \\ \cline{3-13} 
\multicolumn{1}{|l|}{} & \multicolumn{1}{l|}{} & \multicolumn{1}{l|}{\textbf{Scottish Gaelic}} & \multicolumn{1}{r|}{59.7} & \multicolumn{1}{r|}{{\ul 62.7}} & \multicolumn{1}{r|}{{\ul 62.7}} & \multicolumn{1}{r|}{58.2} & \multicolumn{1}{r|}{61.2} & \multicolumn{1}{r|}{47.3} & \multicolumn{1}{r|}{52.2} & \multicolumn{1}{r|}{53.2} & \multicolumn{1}{r|}{58.2} & \multicolumn{1}{r|}{\textbf{64.7}} \\ \cline{3-13} 
\multicolumn{1}{|l|}{} & \multicolumn{1}{l|}{} & \multicolumn{1}{l|}{\textbf{Sinhala}} & \multicolumn{1}{r|}{58.2} & \multicolumn{1}{r|}{55.2} & \multicolumn{1}{r|}{53.7} & \multicolumn{1}{r|}{61.2} & \multicolumn{1}{r|}{61.2} & \multicolumn{1}{r|}{57.7} & \multicolumn{1}{r|}{{\ul 65.7}} & \multicolumn{1}{r|}{54.2} & \multicolumn{1}{r|}{67.7} & \multicolumn{1}{r|}{\textbf{73.6}} \\ \cline{3-13} 
\multicolumn{1}{|l|}{} & \multicolumn{1}{l|}{} & \multicolumn{1}{l|}{\textbf{Swahili}} & \multicolumn{1}{r|}{73.6} & \multicolumn{1}{r|}{71.6} & \multicolumn{1}{r|}{63.2} & \multicolumn{1}{r|}{{\ul \textbf{76.1}}} & \multicolumn{1}{r|}{72.1} & \multicolumn{1}{r|}{57.2} & \multicolumn{1}{r|}{64.7} & \multicolumn{1}{r|}{67.7} & \multicolumn{1}{r|}{72.6} & \multicolumn{1}{r|}{74.6} \\ \cline{3-13} 
\multicolumn{1}{|l|}{} & \multicolumn{1}{l|}{} & \multicolumn{1}{l|}{\textbf{Telugu}} & \multicolumn{1}{r|}{\textbf{86.6}} & \multicolumn{1}{r|}{{\ul 84.6}} & \multicolumn{1}{r|}{83.1} & \multicolumn{1}{r|}{{\ul 84.6}} & \multicolumn{1}{r|}{{\ul 84.6}} & \multicolumn{1}{r|}{84.1} & \multicolumn{1}{r|}{84.1} & \multicolumn{1}{r|}{83.1} & \multicolumn{1}{r|}{82.1} & \multicolumn{1}{r|}{86.1} \\ \cline{3-13} 
\multicolumn{1}{|l|}{} & \multicolumn{1}{l|}{} & \multicolumn{1}{l|}{\textbf{Yoruba}} & \multicolumn{1}{r|}{62.2} & \multicolumn{1}{r|}{59.7} & \multicolumn{1}{r|}{59.7} & \multicolumn{1}{r|}{60.7} & \multicolumn{1}{r|}{60.7} & \multicolumn{1}{r|}{61.7} & \multicolumn{1}{r|}{{\ul 62.2}} & \multicolumn{1}{r|}{{\ul 62.2}} & \multicolumn{1}{r|}{60.7} & \multicolumn{1}{r|}{\textbf{67.2}} \\ \cline{1-1} \cline{3-13} 
 &  &  & \multicolumn{1}{l}{} & \multicolumn{1}{l}{} & \multicolumn{1}{l}{} & \multicolumn{1}{l}{} & \multicolumn{1}{l}{} & \multicolumn{1}{l}{} & \multicolumn{1}{l}{} & \multicolumn{1}{l}{} & \multicolumn{1}{l}{} & \multicolumn{1}{l}{} \\ \cline{1-1} \cline{3-13} 
\multicolumn{1}{|l|}{\multirow{12}{*}{\textbf{Belebele}}} & \multicolumn{1}{l|}{} & \multicolumn{1}{l|}{\textbf{Amharic}} & \multicolumn{1}{r|}{30.0} & \multicolumn{1}{r|}{29.5} & \multicolumn{1}{r|}{28.3} & \multicolumn{1}{r|}{29.9} & \multicolumn{1}{r|}{27.4} & \multicolumn{1}{r|}{30.2} & \multicolumn{1}{r|}{{\ul 31.9}} & \multicolumn{1}{r|}{29.1} & \multicolumn{1}{r|}{30.9} & \multicolumn{1}{r|}{\textbf{36.8}} \\ \cline{3-13} 
\multicolumn{1}{|l|}{} & \multicolumn{1}{l|}{} & \multicolumn{1}{l|}{\textbf{Bengali}} & \multicolumn{1}{r|}{64.5} & \multicolumn{1}{r|}{49.5} & \multicolumn{1}{r|}{48.9} & \multicolumn{1}{r|}{63.9} & \multicolumn{1}{r|}{63.2} & \multicolumn{1}{r|}{58.1} & \multicolumn{1}{r|}{63.9} & \multicolumn{1}{r|}{{\ul 65.6}} & \multicolumn{1}{r|}{64.4} & \multicolumn{1}{r|}{\textbf{65.9}} \\ \cline{3-13} 
\multicolumn{1}{|l|}{} & \multicolumn{1}{l|}{} & \multicolumn{1}{l|}{\textbf{Guarani}} & \multicolumn{1}{r|}{33.3} & \multicolumn{1}{r|}{29.7} & \multicolumn{1}{r|}{30.8} & \multicolumn{1}{r|}{{\ul 35.6}} & \multicolumn{1}{r|}{31.5} & \multicolumn{1}{r|}{34.4} & \multicolumn{1}{r|}{33.6} & \multicolumn{1}{r|}{34.1} & \multicolumn{1}{r|}{32.9} & \multicolumn{1}{r|}{\textbf{40.4}} \\ \cline{3-13} 
\multicolumn{1}{|l|}{} & \multicolumn{1}{l|}{} & \multicolumn{1}{l|}{\textbf{Javanese}} & \multicolumn{1}{r|}{50.8} & \multicolumn{1}{r|}{56.1} & \multicolumn{1}{r|}{56.5} & \multicolumn{1}{r|}{60.4} & \multicolumn{1}{r|}{{\ul 60.9}} & \multicolumn{1}{r|}{48.9} & \multicolumn{1}{r|}{45.8} & \multicolumn{1}{r|}{49.2} & \multicolumn{1}{r|}{48.0} & \multicolumn{1}{r|}{\textbf{65.1}} \\ \cline{3-13} 
\multicolumn{1}{|l|}{} & \multicolumn{1}{l|}{} & \multicolumn{1}{l|}{\textbf{Kyrgyz}} & \multicolumn{1}{r|}{47.4} & \multicolumn{1}{r|}{46.8} & \multicolumn{1}{r|}{47.7} & \multicolumn{1}{r|}{40.8} & \multicolumn{1}{r|}{{\ul 52.8}} & \multicolumn{1}{r|}{52.2} & \multicolumn{1}{r|}{51.7} & \multicolumn{1}{r|}{47.4} & \multicolumn{1}{r|}{47.6} & \multicolumn{1}{r|}{\textbf{56.9}} \\ \cline{3-13} 
\multicolumn{1}{|l|}{} & \multicolumn{1}{l|}{} & \multicolumn{1}{l|}{\textbf{Lao}} & \multicolumn{1}{r|}{30.9} & \multicolumn{1}{r|}{30.1} & \multicolumn{1}{r|}{32.0} & \multicolumn{1}{r|}{33.3} & \multicolumn{1}{r|}{32.9} & \multicolumn{1}{r|}{34.7} & \multicolumn{1}{r|}{{\ul 35.5}} & \multicolumn{1}{r|}{33.9} & \multicolumn{1}{r|}{36.6} & \multicolumn{1}{r|}{\textbf{36.7}} \\ \cline{3-13} 
\multicolumn{1}{|l|}{} & \multicolumn{1}{l|}{} & \multicolumn{1}{l|}{\textbf{Maori}} & \multicolumn{1}{r|}{31.5} & \multicolumn{1}{r|}{31.9} & \multicolumn{1}{r|}{{\ul 33.7}} & \multicolumn{1}{r|}{33.0} & \multicolumn{1}{r|}{30.8} & \multicolumn{1}{r|}{32.1} & \multicolumn{1}{r|}{31.1} & \multicolumn{1}{r|}{32.0} & \multicolumn{1}{r|}{31.8} & \multicolumn{1}{r|}{\textbf{35.0}} \\ \cline{3-13} 
\multicolumn{1}{|l|}{} & \multicolumn{1}{l|}{} & \multicolumn{1}{l|}{\textbf{Mongolian}} & \multicolumn{1}{r|}{38.1} & \multicolumn{1}{r|}{37.1} & \multicolumn{1}{r|}{37.1} & \multicolumn{1}{r|}{{\ul 39.5}} & \multicolumn{1}{r|}{38.6} & \multicolumn{1}{r|}{37.8} & \multicolumn{1}{r|}{36.8} & \multicolumn{1}{r|}{36.3} & \multicolumn{1}{r|}{37.8} & \multicolumn{1}{r|}{\textbf{47.0}} \\ \cline{3-13} 
\multicolumn{1}{|l|}{} & \multicolumn{1}{l|}{} & \multicolumn{1}{l|}{\textbf{Sinhala}} & \multicolumn{1}{r|}{38.7} & \multicolumn{1}{r|}{39.5} & \multicolumn{1}{r|}{39.0} & \multicolumn{1}{r|}{{\ul 44.3}} & \multicolumn{1}{r|}{42.6} & \multicolumn{1}{r|}{39.6} & \multicolumn{1}{r|}{43.7} & \multicolumn{1}{r|}{41.4} & \multicolumn{1}{r|}{46.5} & \multicolumn{1}{r|}{\textbf{52.4}} \\ \cline{3-13} 
\multicolumn{1}{|l|}{} & \multicolumn{1}{l|}{} & \multicolumn{1}{l|}{\textbf{Swahili}} & \multicolumn{1}{r|}{45.2} & \multicolumn{1}{r|}{39.0} & \multicolumn{1}{r|}{40.0} & \multicolumn{1}{r|}{43.9} & \multicolumn{1}{r|}{{\ul 44.4}} & \multicolumn{1}{r|}{38.5} & \multicolumn{1}{r|}{42.1} & \multicolumn{1}{r|}{40.9} & \multicolumn{1}{r|}{40.4} & \multicolumn{1}{r|}{\textbf{54.6}} \\ \cline{3-13} 
\multicolumn{1}{|l|}{} & \multicolumn{1}{l|}{} & \multicolumn{1}{l|}{\textbf{Telugu}} & \multicolumn{1}{r|}{\textbf{61.0}} & \multicolumn{1}{r|}{56.1} & \multicolumn{1}{r|}{46.2} & \multicolumn{1}{r|}{58.6} & \multicolumn{1}{r|}{57.1} & \multicolumn{1}{r|}{{\ul 60.5}} & \multicolumn{1}{r|}{59.6} & \multicolumn{1}{r|}{57.7} & \multicolumn{1}{r|}{59.3} & \multicolumn{1}{r|}{60.8} \\ \cline{3-13} 
\multicolumn{1}{|l|}{} & \multicolumn{1}{l|}{} & \multicolumn{1}{l|}{\textbf{Yoruba}} & \multicolumn{1}{r|}{29.8} & \multicolumn{1}{r|}{32.0} & \multicolumn{1}{r|}{32.2} & \multicolumn{1}{r|}{{\ul 33.6}} & \multicolumn{1}{r|}{31.9} & \multicolumn{1}{r|}{32.3} & \multicolumn{1}{r|}{33.4} & \multicolumn{1}{r|}{33.2} & \multicolumn{1}{r|}{31.9} & \multicolumn{1}{r|}{\textbf{34.3}} \\ \cline{1-1} \cline{3-13} 
 &  &  & \multicolumn{1}{l}{} & \multicolumn{1}{l}{} & \multicolumn{1}{l}{} & \multicolumn{1}{l}{} & \multicolumn{1}{l}{} & \multicolumn{1}{l}{} & \multicolumn{1}{l}{} & \multicolumn{1}{l}{} & \multicolumn{1}{l}{} & \multicolumn{1}{l}{} \\ \cline{1-1} \cline{3-13} 
\multicolumn{1}{|l|}{\multirow{7}{*}{\textbf{\begin{tabular}[c]{@{}l@{}}Global\\ MMLU\end{tabular}}}} & \multicolumn{1}{l|}{} & \multicolumn{1}{l|}{\textbf{Amharic}} & \multicolumn{1}{r|}{27.9} & \multicolumn{1}{r|}{28.0} & \multicolumn{1}{r|}{27.5} & \multicolumn{1}{r|}{28.5} & \multicolumn{1}{r|}{28.0} & \multicolumn{1}{r|}{28.0} & \multicolumn{1}{r|}{28.4} & \multicolumn{1}{r|}{{\ul 28.6}} & \multicolumn{1}{r|}{28.2} & \multicolumn{1}{r|}{\textbf{29.4}} \\ \cline{3-13} 
\multicolumn{1}{|l|}{} & \multicolumn{1}{l|}{} & \multicolumn{1}{l|}{\textbf{Bengali}} & \multicolumn{1}{r|}{40.5} & \multicolumn{1}{r|}{40.4} & \multicolumn{1}{r|}{39.9} & \multicolumn{1}{r|}{38.3} & \multicolumn{1}{r|}{40.8} & \multicolumn{1}{r|}{{\ul \textbf{41.3}}} & \multicolumn{1}{r|}{41.2} & \multicolumn{1}{r|}{38.8} & \multicolumn{1}{r|}{40.0} & \multicolumn{1}{r|}{40.7} \\ \cline{3-13} 
\multicolumn{1}{|l|}{} & \multicolumn{1}{l|}{} & \multicolumn{1}{l|}{\textbf{Kyrgyz}} & \multicolumn{1}{r|}{33.8} & \multicolumn{1}{r|}{37.0} & \multicolumn{1}{r|}{36.5} & \multicolumn{1}{r|}{32.7} & \multicolumn{1}{r|}{37.1} & \multicolumn{1}{r|}{37.0} & \multicolumn{1}{r|}{{\ul \textbf{37.6}}} & \multicolumn{1}{r|}{33.9} & \multicolumn{1}{r|}{33.7} & \multicolumn{1}{r|}{37.5} \\ \cline{3-13} 
\multicolumn{1}{|l|}{} & \multicolumn{1}{l|}{} & \multicolumn{1}{l|}{\textbf{Sinhala}} & \multicolumn{1}{r|}{30.4} & \multicolumn{1}{r|}{30.3} & \multicolumn{1}{r|}{30.5} & \multicolumn{1}{r|}{30.3} & \multicolumn{1}{r|}{30.9} & \multicolumn{1}{r|}{30.4} & \multicolumn{1}{r|}{30.3} & \multicolumn{1}{r|}{{\ul 31.0}} & \multicolumn{1}{r|}{31.0} & \multicolumn{1}{r|}{\textbf{32.3}} \\ \cline{3-13} 
\multicolumn{1}{|l|}{} & \multicolumn{1}{l|}{} & \multicolumn{1}{l|}{\textbf{Swahili}} & \multicolumn{1}{r|}{32.9} & \multicolumn{1}{r|}{31.6} & \multicolumn{1}{r|}{32.5} & \multicolumn{1}{r|}{{\ul 33.3}} & \multicolumn{1}{r|}{32.6} & \multicolumn{1}{r|}{30.1} & \multicolumn{1}{r|}{30.2} & \multicolumn{1}{r|}{31.7} & \multicolumn{1}{r|}{31.2} & \multicolumn{1}{r|}{\textbf{35.1}} \\ \cline{3-13} 
\multicolumn{1}{|l|}{} & \multicolumn{1}{l|}{} & \multicolumn{1}{l|}{\textbf{Telugu}} & \multicolumn{1}{r|}{\textbf{38.9}} & \multicolumn{1}{r|}{{\ul 40.2}} & \multicolumn{1}{r|}{38.9} & \multicolumn{1}{r|}{39.9} & \multicolumn{1}{r|}{39.6} & \multicolumn{1}{r|}{38.6} & \multicolumn{1}{r|}{37.9} & \multicolumn{1}{r|}{39.3} & \multicolumn{1}{r|}{37.7} & \multicolumn{1}{r|}{38.7} \\ \cline{3-13} 
\multicolumn{1}{|l|}{} & \multicolumn{1}{l|}{} & \multicolumn{1}{l|}{\textbf{Yoruba}} & \multicolumn{1}{r|}{31.1} & \multicolumn{1}{r|}{{\ul 33.2}} & \multicolumn{1}{r|}{32.9} & \multicolumn{1}{r|}{31.2} & \multicolumn{1}{r|}{31.9} & \multicolumn{1}{r|}{32.7} & \multicolumn{1}{r|}{32.9} & \multicolumn{1}{r|}{31.1} & \multicolumn{1}{r|}{31.2} & \multicolumn{1}{r|}{\textbf{33.4}} \\ \cline{1-1} \cline{3-13} 
 &  &  & \multicolumn{1}{l}{} & \multicolumn{1}{l}{} & \multicolumn{1}{l}{} & \multicolumn{1}{l}{} & \multicolumn{1}{l}{} & \multicolumn{1}{l}{} & \multicolumn{1}{l}{} & \multicolumn{1}{l}{} & \multicolumn{1}{l}{} & \multicolumn{1}{l}{} \\ \cline{1-1} \cline{3-13} 
\multicolumn{1}{|l|}{\multirow{14}{*}{\textbf{\begin{tabular}[c]{@{}l@{}}FLORES\\ xx-en\end{tabular}}}} & \multicolumn{1}{l|}{} & \multicolumn{1}{l|}{\textbf{Amharic}} & \multicolumn{1}{r|}{18.8} & \multicolumn{1}{r|}{3.2} & \multicolumn{1}{r|}{8.9} & \multicolumn{1}{r|}{17.3} & \multicolumn{1}{r|}{17.1} & \multicolumn{1}{r|}{13.2} & \multicolumn{1}{r|}{17.0} & \multicolumn{1}{r|}{{\ul 17.3}} & \multicolumn{1}{r|}{17.7} & \multicolumn{1}{r|}{\textbf{31.6}} \\ \cline{3-13} 
\multicolumn{1}{|l|}{} & \multicolumn{1}{l|}{} & \multicolumn{1}{l|}{\textbf{Bengali}} & \multicolumn{1}{r|}{\textbf{53.5}} & \multicolumn{1}{r|}{0.7} & \multicolumn{1}{r|}{18.8} & \multicolumn{1}{r|}{49.4} & \multicolumn{1}{r|}{39.6} & \multicolumn{1}{r|}{33.0} & \multicolumn{1}{r|}{49.9} & \multicolumn{1}{r|}{{\ul 52.2}} & \multicolumn{1}{r|}{52.9} & \multicolumn{1}{r|}{51.1} \\ \cline{3-13} 
\multicolumn{1}{|l|}{} & \multicolumn{1}{l|}{} & \multicolumn{1}{l|}{\textbf{Galician}} & \multicolumn{1}{r|}{\textbf{63.5}} & \multicolumn{1}{r|}{62.5} & \multicolumn{1}{r|}{62.6} & \multicolumn{1}{r|}{{\ul 62.7}} & \multicolumn{1}{r|}{62.0} & \multicolumn{1}{r|}{62.8} & \multicolumn{1}{r|}{62.9} & \multicolumn{1}{r|}{{\ul 62.7}} & \multicolumn{1}{r|}{62.5} & \multicolumn{1}{r|}{63.2} \\ \cline{3-13} 
\multicolumn{1}{|l|}{} & \multicolumn{1}{l|}{} & \multicolumn{1}{l|}{\textbf{Guarani}} & \multicolumn{1}{r|}{28.2} & \multicolumn{1}{r|}{28.8} & \multicolumn{1}{r|}{28.3} & \multicolumn{1}{r|}{{\ul 30.4}} & \multicolumn{1}{r|}{28.7} & \multicolumn{1}{r|}{27.7} & \multicolumn{1}{r|}{29.4} & \multicolumn{1}{r|}{{\ul 30.4}} & \multicolumn{1}{r|}{30.6} & \multicolumn{1}{r|}{\textbf{35.4}} \\ \cline{3-13} 
\multicolumn{1}{|l|}{} & \multicolumn{1}{l|}{} & \multicolumn{1}{l|}{\textbf{Javanese}} & \multicolumn{1}{r|}{44.9} & \multicolumn{1}{r|}{47.1} & \multicolumn{1}{r|}{47.7} & \multicolumn{1}{r|}{47.7} & \multicolumn{1}{r|}{{\ul 48.4}} & \multicolumn{1}{r|}{46.6} & \multicolumn{1}{r|}{47.2} & \multicolumn{1}{r|}{47.9} & \multicolumn{1}{r|}{47.9} & \multicolumn{1}{r|}{\textbf{53.4}} \\ \cline{3-13} 
\multicolumn{1}{|l|}{} & \multicolumn{1}{l|}{} & \multicolumn{1}{l|}{\textbf{Kyrgyz}} & \multicolumn{1}{r|}{38.3} & \multicolumn{1}{r|}{37.6} & \multicolumn{1}{r|}{36.9} & \multicolumn{1}{r|}{14.8} & \multicolumn{1}{r|}{29.5} & \multicolumn{1}{r|}{38.9} & \multicolumn{1}{r|}{33.9} & \multicolumn{1}{r|}{{\ul 39.3}} & \multicolumn{1}{r|}{39.4} & \multicolumn{1}{r|}{\textbf{43.1}} \\ \cline{3-13} 
\multicolumn{1}{|l|}{} & \multicolumn{1}{l|}{} & \multicolumn{1}{l|}{\textbf{Lao}} & \multicolumn{1}{r|}{34.1} & \multicolumn{1}{r|}{26.5} & \multicolumn{1}{r|}{26.3} & \multicolumn{1}{r|}{31.2} & \multicolumn{1}{r|}{31.2} & \multicolumn{1}{r|}{29.3} & \multicolumn{1}{r|}{28.7} & \multicolumn{1}{r|}{{\ul 31.5}} & \multicolumn{1}{r|}{33.6} & \multicolumn{1}{r|}{\textbf{38.8}} \\ \cline{3-13} 
\multicolumn{1}{|l|}{} & \multicolumn{1}{l|}{} & \multicolumn{1}{l|}{\textbf{Maori}} & \multicolumn{1}{r|}{27.3} & \multicolumn{1}{r|}{26.6} & \multicolumn{1}{r|}{27.1} & \multicolumn{1}{r|}{29.3} & \multicolumn{1}{r|}{28.0} & \multicolumn{1}{r|}{26.5} & \multicolumn{1}{r|}{28.0} & \multicolumn{1}{r|}{{\ul 29.8}} & \multicolumn{1}{r|}{29.9} & \multicolumn{1}{r|}{\textbf{38.9}} \\ \cline{3-13} 
\multicolumn{1}{|l|}{} & \multicolumn{1}{l|}{} & \multicolumn{1}{l|}{\textbf{Mongolian}} & \multicolumn{1}{r|}{29.4} & \multicolumn{1}{r|}{32.5} & \multicolumn{1}{r|}{32.2} & \multicolumn{1}{r|}{34.1} & \multicolumn{1}{r|}{33.6} & \multicolumn{1}{r|}{31.6} & \multicolumn{1}{r|}{33.7} & \multicolumn{1}{r|}{{\ul 34.7}} & \multicolumn{1}{r|}{34.4} & \multicolumn{1}{r|}{\textbf{42.8}} \\ \cline{3-13} 
\multicolumn{1}{|l|}{} & \multicolumn{1}{l|}{} & \multicolumn{1}{l|}{\textbf{Scottish Gaelic}} & \multicolumn{1}{r|}{31.3} & \multicolumn{1}{r|}{33.6} & \multicolumn{1}{r|}{33.2} & \multicolumn{1}{r|}{{\ul 36.0}} & \multicolumn{1}{r|}{35.2} & \multicolumn{1}{r|}{28.0} & \multicolumn{1}{r|}{29.3} & \multicolumn{1}{r|}{31.6} & \multicolumn{1}{r|}{32.4} & \multicolumn{1}{r|}{\textbf{44.3}} \\ \cline{3-13} 
\multicolumn{1}{|l|}{} & \multicolumn{1}{l|}{} & \multicolumn{1}{l|}{\textbf{Sinhala}} & \multicolumn{1}{r|}{29.3} & \multicolumn{1}{r|}{1.9} & \multicolumn{1}{r|}{2.6} & \multicolumn{1}{r|}{{\ul 31.0}} & \multicolumn{1}{r|}{3.7} & \multicolumn{1}{r|}{5.5} & \multicolumn{1}{r|}{6.2} & \multicolumn{1}{r|}{29.9} & \multicolumn{1}{r|}{33.3} & \multicolumn{1}{r|}{\textbf{41.7}} \\ \cline{3-13} 
\multicolumn{1}{|l|}{} & \multicolumn{1}{l|}{} & \multicolumn{1}{l|}{\textbf{Swahili}} & \multicolumn{1}{r|}{41.0} & \multicolumn{1}{r|}{39.9} & \multicolumn{1}{r|}{39.5} & \multicolumn{1}{r|}{{\ul 42.8}} & \multicolumn{1}{r|}{41.2} & \multicolumn{1}{r|}{36.5} & \multicolumn{1}{r|}{41.2} & \multicolumn{1}{r|}{42.0} & \multicolumn{1}{r|}{42.2} & \multicolumn{1}{r|}{\textbf{52.8}} \\ \cline{3-13} 
\multicolumn{1}{|l|}{} & \multicolumn{1}{l|}{} & \multicolumn{1}{l|}{\textbf{Telugu}} & \multicolumn{1}{r|}{\textbf{54.9}} & \multicolumn{1}{r|}{48.2} & \multicolumn{1}{r|}{49.2} & \multicolumn{1}{r|}{47.3} & \multicolumn{1}{r|}{52.9} & \multicolumn{1}{r|}{48.0} & \multicolumn{1}{r|}{54.8} & \multicolumn{1}{r|}{{\ul \textbf{54.9}}} & \multicolumn{1}{r|}{54.2} & \multicolumn{1}{r|}{54.2} \\ \cline{3-13} 
\multicolumn{1}{|l|}{} & \multicolumn{1}{l|}{} & \multicolumn{1}{l|}{\textbf{Yoruba}} & \multicolumn{1}{r|}{32.5} & \multicolumn{1}{r|}{29.9} & \multicolumn{1}{r|}{29.9} & \multicolumn{1}{r|}{{\ul 33.7}} & \multicolumn{1}{r|}{27.1} & \multicolumn{1}{r|}{31.8} & \multicolumn{1}{r|}{32.3} & \multicolumn{1}{r|}{32.7} & \multicolumn{1}{r|}{33.2} & \multicolumn{1}{r|}{\textbf{36.9}} \\ \cline{1-1} \cline{3-13} 
 &  &  & \multicolumn{1}{l}{} & \multicolumn{1}{l}{} & \multicolumn{1}{l}{} & \multicolumn{1}{l}{} & \multicolumn{1}{l}{} & \multicolumn{1}{l}{} & \multicolumn{1}{l}{} & \multicolumn{1}{l}{} & \multicolumn{1}{l}{} & \multicolumn{1}{l}{} \\ \cline{1-1} \cline{3-13} 
\multicolumn{1}{|l|}{\multirow{14}{*}{\textbf{\begin{tabular}[c]{@{}l@{}}FLORES\\ en-xx\end{tabular}}}} & \multicolumn{1}{l|}{} & \multicolumn{1}{l|}{\textbf{Amharic}} & \multicolumn{1}{r|}{1.1} & \multicolumn{1}{r|}{0.5} & \multicolumn{1}{r|}{0.5} & \multicolumn{1}{r|}{0.5} & \multicolumn{1}{r|}{0.4} & \multicolumn{1}{r|}{{\ul 0.6}} & \multicolumn{1}{r|}{0.3} & \multicolumn{1}{r|}{0.2} & \multicolumn{1}{r|}{0.3} & \multicolumn{1}{r|}{\textbf{8.4}} \\ \cline{3-13} 
\multicolumn{1}{|l|}{} & \multicolumn{1}{l|}{} & \multicolumn{1}{l|}{\textbf{Bengali}} & \multicolumn{1}{r|}{\textbf{37.4}} & \multicolumn{1}{r|}{30.1} & \multicolumn{1}{r|}{31.4} & \multicolumn{1}{r|}{29.8} & \multicolumn{1}{r|}{31.2} & \multicolumn{1}{r|}{{\ul 35.7}} & \multicolumn{1}{r|}{34.7} & \multicolumn{1}{r|}{34.1} & \multicolumn{1}{r|}{35.1} & \multicolumn{1}{r|}{36.5} \\ \cline{3-13} 
\multicolumn{1}{|l|}{} & \multicolumn{1}{l|}{} & \multicolumn{1}{l|}{\textbf{Galician}} & \multicolumn{1}{r|}{52.5} & \multicolumn{1}{r|}{52.1} & \multicolumn{1}{r|}{52.6} & \multicolumn{1}{r|}{52.5} & \multicolumn{1}{r|}{52.7} & \multicolumn{1}{r|}{52.0} & \multicolumn{1}{r|}{{\ul 53.1}} & \multicolumn{1}{r|}{52.6} & \multicolumn{1}{r|}{52.6} & \multicolumn{1}{r|}{\textbf{54.0}} \\ \cline{3-13} 
\multicolumn{1}{|l|}{} & \multicolumn{1}{l|}{} & \multicolumn{1}{l|}{\textbf{Guarani}} & \multicolumn{1}{r|}{5.6} & \multicolumn{1}{r|}{8.0} & \multicolumn{1}{r|}{8.3} & \multicolumn{1}{r|}{9.7} & \multicolumn{1}{r|}{10.4} & \multicolumn{1}{r|}{{\ul 11.7}} & \multicolumn{1}{r|}{7.9} & \multicolumn{1}{r|}{7.6} & \multicolumn{1}{r|}{9.0} & \multicolumn{1}{r|}{\textbf{22.1}} \\ \cline{3-13} 
\multicolumn{1}{|l|}{} & \multicolumn{1}{l|}{} & \multicolumn{1}{l|}{\textbf{Javanese}} & \multicolumn{1}{r|}{27.2} & \multicolumn{1}{r|}{{\ul 34.3}} & \multicolumn{1}{r|}{{\ul 34.3}} & \multicolumn{1}{r|}{32.3} & \multicolumn{1}{r|}{33.1} & \multicolumn{1}{r|}{32.6} & \multicolumn{1}{r|}{32.2} & \multicolumn{1}{r|}{32.8} & \multicolumn{1}{r|}{33.5} & \multicolumn{1}{r|}{\textbf{43.8}} \\ \cline{3-13} 
\multicolumn{1}{|l|}{} & \multicolumn{1}{l|}{} & \multicolumn{1}{l|}{\textbf{Kyrgyz}} & \multicolumn{1}{r|}{18.5} & \multicolumn{1}{r|}{18.7} & \multicolumn{1}{r|}{17.8} & \multicolumn{1}{r|}{17.2} & \multicolumn{1}{r|}{18.7} & \multicolumn{1}{r|}{{\ul 19.6}} & \multicolumn{1}{r|}{18.6} & \multicolumn{1}{r|}{14.9} & \multicolumn{1}{r|}{16.3} & \multicolumn{1}{r|}{\textbf{28.6}} \\ \cline{3-13} 
\multicolumn{1}{|l|}{} & \multicolumn{1}{l|}{} & \multicolumn{1}{l|}{\textbf{Lao}} & \multicolumn{1}{r|}{17.5} & \multicolumn{1}{r|}{5.0} & \multicolumn{1}{r|}{4.7} & \multicolumn{1}{r|}{2.8} & \multicolumn{1}{r|}{4.3} & \multicolumn{1}{r|}{4.3} & \multicolumn{1}{r|}{5.2} & \multicolumn{1}{r|}{{\ul 8.9}} & \multicolumn{1}{r|}{6.1} & \multicolumn{1}{r|}{\textbf{19.4}} \\ \cline{3-13} 
\multicolumn{1}{|l|}{} & \multicolumn{1}{l|}{} & \multicolumn{1}{l|}{\textbf{Maori}} & \multicolumn{1}{r|}{11.4} & \multicolumn{1}{r|}{10.9} & \multicolumn{1}{r|}{12.4} & \multicolumn{1}{r|}{{\ul 14.3}} & \multicolumn{1}{r|}{13.2} & \multicolumn{1}{r|}{8.6} & \multicolumn{1}{r|}{9.0} & \multicolumn{1}{r|}{14.0} & \multicolumn{1}{r|}{14.9} & \multicolumn{1}{r|}{\textbf{36.7}} \\ \cline{3-13} 
\multicolumn{1}{|l|}{} & \multicolumn{1}{l|}{} & \multicolumn{1}{l|}{\textbf{Mongolian}} & \multicolumn{1}{r|}{8.7} & \multicolumn{1}{r|}{6.7} & \multicolumn{1}{r|}{6.1} & \multicolumn{1}{r|}{6.2} & \multicolumn{1}{r|}{6.7} & \multicolumn{1}{r|}{{\ul 9.3}} & \multicolumn{1}{r|}{7.3} & \multicolumn{1}{r|}{7.6} & \multicolumn{1}{r|}{7.7} & \multicolumn{1}{r|}{\textbf{22.6}} \\ \cline{3-13} 
\multicolumn{1}{|l|}{} & \multicolumn{1}{l|}{} & \multicolumn{1}{l|}{\textbf{Scottish Gaelic}} & \multicolumn{1}{r|}{10.4} & \multicolumn{1}{r|}{13.1} & \multicolumn{1}{r|}{10.6} & \multicolumn{1}{r|}{{\ul 12.6}} & \multicolumn{1}{r|}{11.1} & \multicolumn{1}{r|}{5.5} & \multicolumn{1}{r|}{5.9} & \multicolumn{1}{r|}{5.5} & \multicolumn{1}{r|}{7.6} & \multicolumn{1}{r|}{\textbf{32.3}} \\ \cline{3-13} 
\multicolumn{1}{|l|}{} & \multicolumn{1}{l|}{} & \multicolumn{1}{l|}{\textbf{Sinhala}} & \multicolumn{1}{r|}{13.2} & \multicolumn{1}{r|}{8.7} & \multicolumn{1}{r|}{9.2} & \multicolumn{1}{r|}{6.6} & \multicolumn{1}{r|}{9.2} & \multicolumn{1}{r|}{8.1} & \multicolumn{1}{r|}{{\ul 10.1}} & \multicolumn{1}{r|}{7.7} & \multicolumn{1}{r|}{12.4} & \multicolumn{1}{r|}{\textbf{22.2}} \\ \cline{3-13} 
\multicolumn{1}{|l|}{} & \multicolumn{1}{l|}{} & \multicolumn{1}{l|}{\textbf{Swahili}} & \multicolumn{1}{r|}{20.1} & \multicolumn{1}{r|}{15.7} & \multicolumn{1}{r|}{14.9} & \multicolumn{1}{r|}{14.7} & \multicolumn{1}{r|}{{\ul 16.0}} & \multicolumn{1}{r|}{10.6} & \multicolumn{1}{r|}{14.8} & \multicolumn{1}{r|}{14.6} & \multicolumn{1}{r|}{12.6} & \multicolumn{1}{r|}{\textbf{38.8}} \\ \cline{3-13} 
\multicolumn{1}{|l|}{} & \multicolumn{1}{l|}{} & \multicolumn{1}{l|}{\textbf{Telugu}} & \multicolumn{1}{r|}{40.2} & \multicolumn{1}{r|}{35.9} & \multicolumn{1}{r|}{36.4} & \multicolumn{1}{r|}{35.5} & \multicolumn{1}{r|}{36.2} & \multicolumn{1}{r|}{{\ul 39.1}} & \multicolumn{1}{r|}{38.4} & \multicolumn{1}{r|}{38.2} & \multicolumn{1}{r|}{39.4} & \multicolumn{1}{r|}{\textbf{40.3}} \\ \cline{3-13} 
\multicolumn{1}{|l|}{} & \multicolumn{1}{l|}{} & \multicolumn{1}{l|}{\textbf{Yoruba}} & \multicolumn{1}{r|}{15.7} & \multicolumn{1}{r|}{19.5} & \multicolumn{1}{r|}{19.2} & \multicolumn{1}{r|}{18.6} & \multicolumn{1}{r|}{{\ul 19.6}} & \multicolumn{1}{r|}{18.0} & \multicolumn{1}{r|}{{\ul 19.6}} & \multicolumn{1}{r|}{16.6} & \multicolumn{1}{r|}{18.0} & \multicolumn{1}{r|}{\textbf{23.9}} \\ \cline{1-1} \cline{3-13} 
\end{tabular}%
}
\caption{Full results of all trained models on their relevant tasks for our ``in-depth'' languages. Values are percentage accuracy for SIB200, Belebele, and GlobalMMLU, and CHRF++ for FLORES tasks. \textbf{Bolded} is the highest value in each row and {\ul{underlined}} is the highest value within the models whose training data was kept constant.}
\label{tab:full_results_table}
\end{table*}

\subsection{Data subsets}

We ran the data generation pipeline, as described in Section~\ref{generatingdata}, with each of the above languages. We then trained student models on the following subsets of data as a comparison between training datasets:

\begin{itemize}[noitemsep,topsep=5pt,parsep=0pt,partopsep=0pt]
    \item \textbf{Topic} - Topic prompt generated data only 
    \item \textbf{Scen} - Scenario prompt generated data only
    \item \textbf{Cont} - Context prompt generated data only
    \item \textbf{Gen} - All generated data (Topic+Scen+Cont)
    \item \textbf{Tran} - Translated data only
    \item \textbf{Gen+Tran} - Generated and translated data 
    \item \textbf{GenReas + Tran (tok-lim)} - Generated data with reasoning traces and translation data, limited at the same number of tokens as all above datasets
    \item \textbf{GenReas + Tran (ex-lim)} - Generated data with reasoning traces and translation data, with the number of examples and non-reasoning tokens limited to the same as Gen+Tran, but with added reasoning tokens
\end{itemize}

We counted the number of tokens of each subset, found the minimum per language, and then sampled the subsets to create seven comparable subsets with the same number of tokens for each language. Over all languages, this averaged to 6.5 million tokens per subset, with the lowest token count being 3.2 million (Galician) and the highest being 14.9 million (Telugu).

Additionally, with the \textbf{Gen (w. Reasoning) + Tran (example limited)} model, we compared other models against the practical situation where reasoning traces were freely available after generating our synthetic data. In this case, we kept the number of tokens and examples from our generated data equal to other models, but added reasoning traces on top of other data. This resulted in datasets that had between 5-11\% more tokens than their token limited counterparts. This was practical in situations such as ours, where data generation costs are more expensive than training costs.

Finally, we trained our full model on all the available data without limiting token count, which we called ``\textbf{GenReas+Tran (full)}''. This resulted in datasets of between 23 and 77 million tokens, or between 34 and 50 thousand unique examples.

\subsection{Evaluation benchmarks}
\label{sec:eval_benchmarks}

We evaluated our models on 4 diverse multilingual benchmark datasets, Belebele~\cite{bandarkar2024belebele}, FLORES~\cite{nllb-24}, GlobalMMLU~\cite{singh2025global}, and SIB200~\cite{adelani2024sib}. These datasets offer good coverage over our 14 languages and cover 5 distinct tasks.

These tasks cover 5 important use-cases of a generalist low-resource language SLM: translation from English to that language, translation to English from that language, classification of text, open question answering, and closed question answering.

The evaluations on these tasks were done with a 3-shot learning template, where the first 3 instances of each dataset are removed from the test set and used as example responses. The temperature used was 0 and the repetition penalty was set to 1.0 to prevent the log-probabilities of candidate answer tokens (e.g., in single token multiple-choice tasks) from being penalized simply because they appeared in the few-shot examples.

We evaluate using accuracy for Belebele, GlobalMMLU, and SIB200, and CHRF++~\cite{popovic2015chrf} for the FLORES tasks. All evaluations used ``non-thinking mode'' (i.e. no reasoning trace is generated).

\subsection{Manual evaluation}

Further to our automatic analysis, we also conducted a manual preference analysis between the outputs of both our final trained Kakugo model (denoted "\textbf{Gen (w. Reasoning) + Tran (full)}" in our other evaluation) and the base student model (Granite 4 Micro) to conversational prompts in three low resource languages: Galician, Javanese, and Yoruba.

We first took the 80 initial prompts from the MT-Bench conversation benchmark~\cite{zheng2023judging} and translated them using GPT-5~\cite{gpt5} into the three languages. Then a native speaker of each of the chosen languages (all authors of this paper) corrected any incorrect or unnatural translations. These prompts were then input to both the Kakugo model and base student model with a temperature of 0.0 (as recommended by the Granite documentation) and the repetition penalty set to 1.05 to prevent repetition of words when outputting low resource language script.

These responses were randomly shuffled, thus making them double-blind, and the native speaker chose which of the two responses best answered the prompt, with the additional option to instead say that the responses were of equal quality.

\section{Results}

\begin{table}[]
\centering
\begin{tabular}{|ll|ll|ll|ll|}
\hline
\textbf{am} & \textbf{5/5} & \textbf{gd} & \textbf{3/3} & \textbf{mi} & \textbf{4/4} & \textbf{su} & \textbf{4/4} \\ \hline
\textbf{arz} & \textbf{4/4} & \textbf{gl} & 2/3 & \textbf{mn} & \textbf{4/4} & \textbf{sw} & \textbf{5/5} \\ \hline
\textbf{ars} & 2/4 & \textbf{gn} & \textbf{4/4} & \textbf{mt} & \textbf{4/4} & \textbf{te} & 1/5 \\ \hline
\textbf{as} & \textbf{4/4} & \textbf{ha} & \textbf{5/5} & \textbf{ny} & \textbf{4/4} & \textbf{tg} & \textbf{4/4} \\ \hline
\textbf{ast} & \textbf{3/3} & \textbf{ht} & \textbf{4/4} & \textbf{oc} & 2/3 & \textbf{ti} & \textbf{4/4} \\ \hline
\textbf{az} & 1/3 & \textbf{ig} & \textbf{5/5} & \textbf{pap} & \textbf{3/3} & \textbf{tk} & \textbf{3/3} \\ \hline
\textbf{ba} & \textbf{3/3} & \textbf{jv} & \textbf{4/4} & \textbf{ps} & \textbf{4/4} & \textbf{tt} & \textbf{3/3} \\ \hline
\textbf{bn} & 2/5 & \textbf{kmr} & \textbf{3/3} & \textbf{rn} & \textbf{3/3} & \textbf{ug} & \textbf{3/3} \\ \hline
\textbf{bo} & 2/4 & \textbf{sdh} & \textbf{4/4} & \textbf{rw} & \textbf{4/4} & \textbf{xh} & \textbf{4/4} \\ \hline
\textbf{ceb} & \textbf{4/4} & \textbf{ky} & \textbf{5/5} & \textbf{sd} & \textbf{4/4} & \textbf{yi} & \textbf{3/3} \\ \hline
\textbf{cv} & 0/2 & \textbf{lb} & 2/3 & \textbf{si} & \textbf{5/5} & \textbf{yo} & \textbf{5/5} \\ \hline
\textbf{cy} & \textbf{3/3} & \textbf{lo} & \textbf{4/4} & \textbf{sm} & \textbf{3/3} & \textbf{zu} & \textbf{4/4} \\ \hline
\textbf{fo} & \textbf{3/3} & \textbf{lus} & 2/3 & \textbf{sn} & \textbf{5/5} & \textbf{} &  \\ \hline
\textbf{ga} & \textbf{3/3} & \textbf{mg} & \textbf{4/4} & \textbf{st} & \textbf{4/4} & \textbf{} &  \\ \hline
\end{tabular}%

\caption{Win rates of our full Kakugo models (i.e. \textbf{GenReas + Tran (full)}) against the base model (i.e. Granite 4 Micro) for benchmarks in all languages. Bolded are values where the Kakugo model outperforms the base model across all benchmarks for that language.}
\label{tab:win_rates}
\end{table}

Average scores across our 14 in-depth languages are in Table~\ref{tab:full-task-model-comparison} and full results are in Table~\ref{tab:full_results_table}.

When comparing each single method of data creation (i.e. \textbf{Topic only}, \textbf{Scenario only}, \textbf{Context only}, and \textbf{Translated only}), we find that different subsets lead to better performance on different tasks and languages. Since our goal is to create a generalist SLM that can perform many tasks, we find that training data created using multiple data generation methods to be the most appropriate for our Kakugo pipeline.

Comparing \textbf{Gen + Tran} to \textbf{GenReas + Tran (tok-lim)} shows no consistent improvement when reasoning traces replace target-language tokens. However, adding traces on top of target language data (\textbf{GenReas + Tran (ex-lim)}) provides small performance gains. Therefore, we find that developers should  prioritize target-language tokens when under training budget constraints, but leverage reasoning traces if they are available as generation by-products.

Summarized evaluation results of all languages can be found in Table~\ref{tab:win_rates} and full results of the base model and our full model for all languages can be found in Appendix~\ref{sec:base_vs_kakugo}.

Overall, we find that the \textbf{Gen (w. Reasoning) + Tran (full)} model achieves an accuracy increase of 9.9\% on Belebele, 1.9\% on GlobalMMLU, 8.8\% on SIB200, and 8.5 CHRF++ on FLORES (xx-en) and 13.1 CHRF++ on FLORES (en-xx), averaged across our 54 languages compared to the base model. This indicates that our pipeline is effective at improving performance across a diverse set of tasks, suggesting applicability as a generalist, multi-task SLM in low resource languages.

We also find performance jumps from most token-limited to full models, suggesting insufficient training data  in token-limited settings. This leaves the Pareto-optimal data size for low-resource SLM distillation an open question.

Performance gains vary by language. While languages such as Javanese and Yoruba improved significantly, the base model achieved better evaluation results for languages like Telugu and Bengali. We find an inverse relationship between initial base model performance and improvements to the base made by our Kakugo pipeline, Pearson correlations reveal moderate to high anti-correlation across four of five tasks (0.04, -0.60, -0.86, -0.37, -0.38 for Belebele, GlobalMMLU, SIB200, FLORES xx-en, and FLORES en-xx, respectively). Consequently, we find Kakugo is most effective on languages where existing SLMs struggle but may yield diminishing returns where the student model already performs well.

\begin{figure}
    \centering
    \includegraphics[width=\linewidth]{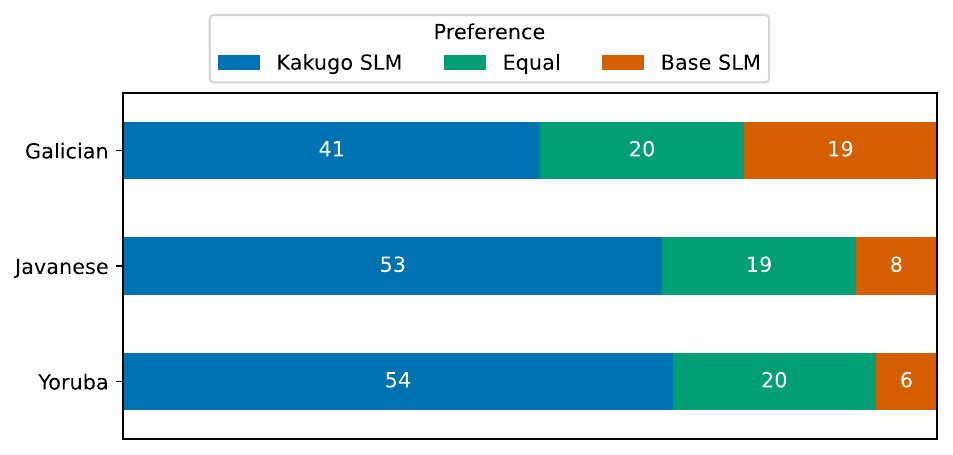}
    \caption{Response preferences from native speakers between responses from our Kakugo model and the base model, Granite 4 Micro.}
    \label{fig:mtbench}
\end{figure}

The results of the manual preference analysis between responses from our Kakugo model and the base student model can be found in Figure~\ref{fig:mtbench}.

We find a consistent preference for most responses from our final Kakugo SLM compared to the base model. This suggests that our Kakugo SLM should be used in conversational settings in low resource languages instead of its base model counterpart. This further indicates the efficacy of our training pipeline for improving the conversational and instruction following abilities of SLMs in low resource languages.

\section{Discussion}

The evaluation results demonstrate that our automated Kakugo data and training pipeline is capable of improving an SLM's low-resource language capabilities. We can see that generated, translated, and reasoning trace data play a key role in improving performance on down-stream tasks, indicating the importance of diversity in data distillation of low resource languages. This is instructive to future research because it shows that even when data seeds are diverse, a single data generation pipeline will not suffice for creating generally-useful SLMs in low-resource languages.

We also find our pipeline to be notably affordable, as we estimate the unit cost of this entire pipeline to be less than \$50. We average between 10-50 million tokens generated per language at a price of \$0.6 per million output tokens\footnote{The current price of our teacher model, GPT-OSS 120B on \href{https://www.together.ai/models/gpt-oss-120b}{TogetherAI}}, and training takes between 1-5 hours on a system priced at \$1.3 per hour\footnote{The approximate current price of a system comparable to which we conducted training on (8 x RTX3090 GPUs) on \href{https://cloud.vast.ai/}{VastAI}}, so even allowing for some redundancy, this pipeline should not exceed \$50 per language. The size of our student models ($\sim$3 billion parameters) means that the cost of evaluation is negligible compared to training.

This affordability gives low-resource language communities access to the benefits of SLMs, such as cheaper, faster, greener, and more data secure inference, without expensive development teams or training costs.

\section{Future work}

Future uses of this technology could include pairing a trained SLM with highly multilingual automatic speech recognition~\cite{keren2025omnilingual,chen2024towards,zhao2025scaling} and text-to-speech technology~\cite{pratap2024scaling} to allow for a fully aural interface for this language model. This could be useful as most natural languages in the world are primarily spoken rather than written~\cite{lewis2010assessing}, meaning a spoken interface may be more natural to speakers of that language.

Another potential use-case of this model is to be used as a small, fast pilot language model in a speculative decoding system~\cite{leviathan2023fast}. This would allow low-resource language outputs to be generated faster with less memory used. This is a particular issue due to the relatively low efficiency of tokenization of low-resource languages~\cite{rust2021good}, often making generation of the same sentence or paragraph slower and more expensive in these languages compared to high-resource languages.

\section{Conclusion}

The Kakugo pipeline shows that a diverse mixture of synthetic reasoning traces and translated data significantly enhances the capabilities of SLMs in low-resource settings. We have open-sourced the resulting models and conversational datasets for 54 languages, many of which previously lacked such resources. Ultimately, this work provides a scalable, affordable framework that empowers low-resource language communities to create efficient, data-sovereign AI models without requiring expensive specialist teams.

\section*{Limitations}

One of the limitations of our experiments is that our results may not generalize beyond the models that we used as the teacher (GPT-OSS 120B) and student (Granite 4 Micro). We chose these models due to their superior base model performance on low-resource language tasks, making these realistic and practical bases from which we train. Instead of repeating our experiments over multiple models, we instead sought to maximize our coverage of different languages, covering more than 50 in this work. Further, our pipeline was not specifically designed for these models and so can theoretically be used with other models too. Future work should replicate our work using other, possibly future, models.

A fundamental limitation of our work is that this pipeline will not work for languages that the teacher model is not proficient at generating data for. For example, if a teacher model has no knowledge of a language, such as Kalamang~\cite{tanzer2023benchmark}, then our automated pipeline approach will not generate any useful training data. Therefore, this approach is restricted to the broad variety of languages which have some presence online or in LLMs but remain low-resource languages that SLMs struggle with.

An area that was not explored extensively in this work is the degree to which datasets should be mixed to maximize down-stream performance. Although we conduct experiments with four separate methods of creating training data (3 prompt generation methods plus translation) and evaluate performance when this data is mixed, it remains an open question as to what mixture of data would lead to the highest evaluation performance. With recent work in evaluating dynamic mixtures of datasets~\cite{wu2024mixture}, future work could investigate exactly which proportions of each type of data evaluated is most appropriate for making a general training pipeline.

Another limitation of this study is that while our evaluations measure the linguistic abilities of language models in low resource languages, they do not explicitly evaluate on topics specific to language or culture. This is because most massively multilingual evaluation benchmarks (such as Belebele, GlobalMMLU, and FLORES) do not have the scope to measure this information, so future work should explore evaluating these models on more culture-specific dimensions.

\section*{Ethical considerations}

While our Kakugo pipeline democratizes access to language technologies by reducing costs and technical barriers, reliance on synthetic and translated data introduces risks of propagating ``translationese'' and foreign cultural norms, potentially erasing unique linguistic nuances \cite{riley2020translationese}. We recognize that while the use of language models has been shown to be positive in many low-resource language settings~\cite{ouattara2025bridging}, not all communities desire the digitization or automation of their language, and some may view non-participatory modeling as a form of extractive digital colonialism \cite{bird2020decolonising, blodgett2020language}. Consequently, this pipeline should be used in collaboration with the target language community whenever possible.

Another ethical consideration is the risk that there is personally identifiable information or offensive content within our synthetically generated data. We took steps to mitigate this risk by choosing a teacher model, GPT-OSS 120B, which has state-of-the-art abilities to refuse to output hateful, illicit, sexual, or personal data in its output~\cite{openai2025gptoss120bgptoss20bmodel} and we did not prompt our teacher model to output this kind of information. However, there is still the risk that this information is contained within our synthetic datasets and therefore this should be considered when using our released datasets.

% \section*{Acknowledgments}
% todo - add Acknowledgments
% Mention HPLT grant

\bibliography{custom}

\newpage

\appendix

\section{Data generation seeds and system prompts}
\label{sec:seeds_and_sys_prompts}

\lstset{
    basicstyle=\ttfamily\small, % Smaller font helps fit more code
    breaklines=true,            % Enable line wrapping
    breakatwhitespace=false,    % Wrap even if there is no space
    frame=single,               % Adds a box to see the column boundary
    columns=fullflexible,        % Improves character spacing
    breakindent=0pt, 
    breakautoindent=false,
}

\subsection{Topic-based prompt seeds}
% \usepackage{graphicx}
\begin{table}[]
\centering
\resizebox{\columnwidth}{!}{%
\begin{tabular}{|l|l|}
\hline
\textbf{General seeds} & \textbf{Language specific seeds} \\ \hline
daily life & \{$language\_name$\} daily life \\ \hline
the world & \{$language\_name$\} culture \\ \hline
health & \{$language\_name$\} health \\ \hline
practical skills & \{$language\_name$\} speaking places \\ \hline
arts and culture & \{$language\_name$\} speaking people \\ \hline
sciences & \{$language\_name$\} language \\ \hline
social sciences & \{$language\_name$\} history \\ \hline
humanities & \{$language\_name$\} society \\ \hline
\end{tabular}%
}
\caption{Seed topics for generating topic-based prompts}
\label{tab:topic_seeds}
\end{table}
See table~\ref{tab:topic_seeds} for lists of general and language specific seeds.

\subsection{Topic generation input}

\begin{lstlisting}
Write {num_topics} topics, things, people, places, objects, concepts, themes, or anything else that encompass the main aspects of {macrotopic}.
Your topics should be diverse, comprehensive, concise, and be written in simple English.
Your topics should be relevant to people who speak {lang_name}.
\end{lstlisting}

\subsection{Topic-based prompt generation input}

\begin{lstlisting}
Write at least {min_prompts} and at most {num_prompts} {lang_name} prompts related to {topic}.
\end{lstlisting}

\subsection{General scenario generation input}

\begin{lstlisting}
Write the {num_scenarios} most popular scenarios that people would use an AI chatbot for.

Do not include asterisks (*) in any of your output.

Your scenarios should be general and concise, rather than detailed.
Your scenarios should be diverse, realistic, practical, comprehensive, concise, and be written in English.

Consider the scenarios in which a {lang_name} speaker would use an AI assistant.
Your scenarios should be specific to users who speak {lang_name}.
\end{lstlisting}

\subsection{Specific scenario generation input}

\begin{lstlisting}
Write {num_scenarios} specific practical scenarios that people would use an AI assistant for when they are using the assistant for the following task: {general_scenario}.

Do not include asterisks (*) in any of your output.

Your scenarios should be general and concise, rather than detailed.
Your scenarios should be diverse, realistic, practical, comprehensive, concise, and be written in English.

Consider the scenarios in which a {lang_name} speaker would use an AI assistant.
Your scenarios should be specific to users who speak {lang_name}.
\end{lstlisting}

\subsection{Scenario-based prompt generation input}

\begin{lstlisting}
Write {n_prompts} prompts in {lang_name} that a user would ask an AI chatbot when they are using the chatbot in the following scenario: {scenario}.
\end{lstlisting}

\subsection{Context-based prompt generation input}

\begin{lstlisting}
Write {num_prompts} prompts in {lang_name} related to the following text:

{context_text}

The prompts should be written in natural {lang_name} and should ask the AI assistant to {prompt_type} the text.
\end{lstlisting}

\subsection{Prompt generation system message}

This system message is used for generating prompts for all of topic, scenario, and context based prompts.

\begin{lstlisting}
You are a {lang_name} speaking user of an AI assistant and you speak in natural {lang_name}.
Your goal is to write a list of prompts in {lang_name}.
The prompts should be diverse, practical, and realistic.
Be as natural and casual as a {lang_name} speaker would be when speaking to a {lang_name} AI assistant.
Do not refer to the assistant directly - be simple where possible.
Use colloquialisms and informal or casual {lang_name} where appropriate to make your prompts as natural and realistic as possible.
Your output should only consist of a single JSON object within a JSON block (delimited by ```json and ```) which has one key, "prompts", which is an array of the {lang_name} prompts.
\end{lstlisting}

\subsection{Prompt revision system message}

\begin{lstlisting}
You are a prompt improvement assistant. Given a prompt in {lang_name}, revise it to make it better.
Your can do any of the following:
* Make the language of the prompt more natural and casual, like how a {lang_name} would actually talk to a helpful assistant.
* Add some context to the original prompt. The context might state the importance of the prompt, explain background knowledge, or add other reasonable information.
* Change the prompt into a different format or style. For example: imperative statements (i.e. adding stipulations for how or how not to output the response), length requirements for the answer (e.g. answer in a single phrase, answer in k words etc.), requiring output in a certain format.
* Elongate prompts that require to elaborate on specific topic or discuss a certain point.
* Add other related questions, statements, or instructions.
Your output should only consist of a single JSON object within a JSON block (delimited by ```json and ```) which has one key, "improved_prompt", which is a string of the improved prompt written in {lang_name}.
\end{lstlisting}

\subsection{Response generation system message}

\begin{lstlisting}
You are a helpful chatbot assistant that speaks in fluent {lang_name}.
Your response should be fully understandable to a {lang_name} speaker.
Speak like a native {lang_name} speaker to the user unless prompted otherwise.
If you give information, always give only correct information.
If you are given instructions, always follow the instructions correctly and fully.
Your response should have minimal formatting where possible - avoid overly complex lists, emphasis, or Markdown formatting unless specified by the user.
If prompted to output a URL, say that you cannot output URLs directly but you can advise ways to find the correct URL (e.g. by suggesting web search terms).
Write your response in as simple and as conversational a tone as possible.
Do not include URLs in your output as they may be hallucinated.
Finally, your responses should be as short as possible - respond with the least words while answering the user's prompt.
\end{lstlisting}

\section{Translation system message}
\label{sec:translation_sys_msg}

\begin{lstlisting}
You are an English to {lang_name} translation assistant.
Given a list of dicts that form a conversation in English between a user and assistant, output the {lang_name} translation of that conversation in the same list of dicts format.
Only include this list of dicts in your final output.
Keep the keys and format of the list of dicts the same in your translation.
\end{lstlisting}

\section{Training hyperparameters}
\label{sec:training_hyperparams}

This is a list of the main training hyperparameters used to train with Llama Factory.

\begin{lstlisting}
model_name_or_path: ibm-granite/granite-4.0-micro
stage: sft
do_train: true
finetuning_type: full
deepspeed: examples/deepspeed/ds_z3_config.json 
template: granite4
cutoff_len: 8000
packing: true
per_device_train_batch_size: 1
gradient_accumulation_steps: 1
learning_rate: 1.0e-5
num_train_epochs: 1.0
lr_scheduler_type: cosine
warmup_ratio: 0.05
bf16: true
val_size: 0.02
per_device_eval_batch_size: 1
eval_strategy: steps
eval_steps: 0.2
\end{lstlisting}

\section{Full results for base vs Kakugo}
\label{sec:base_vs_kakugo}

% \usepackage{graphicx}
\begin{table*}[]
\centering
\resizebox{\textwidth}{!}{%
\begin{tabular}{|l|rr|rr|rr|rr|rr|}
\hline
 & \multicolumn{2}{l|}{\textbf{SIB200}} & \multicolumn{2}{l|}{\textbf{Belebele}} & \multicolumn{2}{l|}{\textbf{GlobalMMLU}} & \multicolumn{2}{l|}{\textbf{FLORES xx-en}} & \multicolumn{2}{l|}{\textbf{FLORES en-xx}} \\ \hline
 & \multicolumn{1}{l}{\textbf{Base}} & \multicolumn{1}{l|}{\textbf{Kakugo}} & \multicolumn{1}{l}{\textbf{Base}} & \multicolumn{1}{l|}{\textbf{Kakugo}} & \multicolumn{1}{l}{\textbf{Base}} & \multicolumn{1}{l|}{\textbf{Kakugo}} & \multicolumn{1}{l}{\textbf{Base}} & \multicolumn{1}{l|}{\textbf{Kakugo}} & \multicolumn{1}{l}{\textbf{Base}} & \multicolumn{1}{l|}{\textbf{Kakugo}} \\ \hline
\textbf{Amharic} & 26.4 & \textbf{51.2} & 30.0 & \textbf{36.8} & 27.9 & \textbf{29.4} & 18.8 & \textbf{31.6} & 1.1 & \textbf{8.4} \\ \hline
\textbf{Aranese} & \textbf{84.6} & 83.6 & - & - & - & - & 59.3 & \textbf{60.1} & 34.1 & \textbf{40.6} \\ \hline
\textbf{Assamese} & 81.1 & \textbf{85.1} & 51.4 & \textbf{58.6} & - & - & 45.2 & \textbf{47.6} & 22.9 & \textbf{28.5} \\ \hline
\textbf{Asturian} & 85.1 & \textbf{85.6} & - & - & - & - & 60.7 & \textbf{61.0} & 42.5 & \textbf{50.3} \\ \hline
\textbf{Bashkir} & 78.1 & \textbf{80.6} & - & - & - & - & 38.2 & \textbf{45.8} & 8.1 & \textbf{24.6} \\ \hline
\textbf{Bengali} & \textbf{85.6} & \textbf{85.6} & 64.5 & \textbf{65.9} & 40.5 & \textbf{40.7} & \textbf{53.5} & 51.1 & \textbf{37.4} & 36.5 \\ \hline
\textbf{Cebuano} & 77.1 & \textbf{80.6} & 45.9 & \textbf{59.2} & - & - & 45.8 & \textbf{57.8} & 32.7 & \textbf{48.8} \\ \hline
\textbf{Central Kurdish} & 64.2 & \textbf{78.1} & 35.1 & \textbf{45.4} & - & - & 31.1 & \textbf{41.3} & 8.9 & \textbf{22.5} \\ \hline
\textbf{Chuvash} & - & - & - & - & - & - & \textbf{26.6} & 23.8 & \textbf{7.2} & 3.1 \\ \hline
\textbf{Eastern Yiddish} & 60.2 & \textbf{76.6} & - & - & - & - & 46.7 & \textbf{61.8} & 16.5 & \textbf{29.7} \\ \hline
\textbf{Egyptian Arabic} & 80.6 & \textbf{82.1} & 54.2 & \textbf{68.0} & - & - & 51.2 & \textbf{52.8} & 34.0 & \textbf{36.2} \\ \hline
\textbf{Faroese} & 72.6 & \textbf{78.6} & - & - & - & - & 43.3 & \textbf{51.1} & 22.0 & \textbf{35.3} \\ \hline
\textbf{Galician} & 85.6 & \textbf{87.1} & - & - & - & - & \textbf{63.5} & 63.2 & 52.5 & \textbf{54.0} \\ \hline
\textbf{Guarani} & 69.2 & \textbf{77.1} & 33.3 & \textbf{40.4} & - & - & 28.2 & \textbf{35.4} & 5.6 & \textbf{22.1} \\ \hline
\textbf{Haitian Creole} & 73.6 & \textbf{83.6} & 42.8 & \textbf{61.4} & - & - & 42.3 & \textbf{55.8} & 15.3 & \textbf{44.7} \\ \hline
\textbf{Hausa} & 62.2 & \textbf{72.6} & 33.7 & \textbf{42.3} & 31.1 & \textbf{33.2} & 34.0 & \textbf{45.0} & 14.0 & \textbf{35.7} \\ \hline
\textbf{Igbo} & 68.2 & \textbf{75.1} & 32.7 & \textbf{39.9} & 32.5 & \textbf{35.1} & 34.3 & \textbf{40.9} & 20.1 & \textbf{31.2} \\ \hline
\textbf{Irish} & 67.2 & \textbf{73.6} & - & - & - & - & 36.6 & \textbf{48.7} & 16.8 & \textbf{29.6} \\ \hline
\textbf{Javanese} & 72.6 & \textbf{83.1} & 50.8 & \textbf{65.1} & - & - & 44.9 & \textbf{53.4} & 27.2 & \textbf{43.8} \\ \hline
\textbf{Kinyarwanda} & 51.2 & \textbf{73.6} & 31.7 & \textbf{48.7} & - & - & 26.3 & \textbf{43.5} & 5.9 & \textbf{27.1} \\ \hline
\textbf{Kyrgyz} & 75.1 & \textbf{81.1} & 47.4 & \textbf{56.9} & 33.8 & \textbf{37.5} & 38.3 & \textbf{43.1} & 18.5 & \textbf{28.6} \\ \hline
\textbf{Lao} & 63.2 & \textbf{67.7} & 30.9 & \textbf{36.7} & - & - & 34.1 & \textbf{38.8} & 17.5 & \textbf{19.4} \\ \hline
\textbf{Lhasa Tibetan} & 27.9 & \textbf{47.3} & \textbf{27.6} & \textbf{27.6} & - & - & 18.4 & \textbf{19.4} & \textbf{0.6} & 0.1 \\ \hline
\textbf{Luxembourgish} & \textbf{83.1} & \textbf{83.1} & - & - & - & - & 56.9 & \textbf{62.2} & 27.4 & \textbf{44.4} \\ \hline
\textbf{Maltese} & 81.1 & \textbf{88.1} & 41.8 & \textbf{61.5} & - & - & 50.4 & \textbf{64.7} & 14.5 & \textbf{41.1} \\ \hline
\textbf{Maori} & 49.3 & \textbf{68.2} & 31.5 & \textbf{35.0} & - & - & 27.3 & \textbf{38.9} & 11.4 & \textbf{36.7} \\ \hline
\textbf{Mizo} & 65.7 & \textbf{69.7} & - & - & - & - & 27.8 & \textbf{28.4} & \textbf{12.5} & 5.2 \\ \hline
\textbf{Mongolian} & 63.2 & \textbf{81.1} & 38.1 & \textbf{47.0} & - & - & 29.4 & \textbf{42.8} & 8.7 & \textbf{22.6} \\ \hline
\textbf{Najdi Arabic} & 83.1 & \textbf{84.1} & 59.3 & \textbf{67.3} & - & - & \textbf{58.7} & 57.4 & \textbf{42.9} & 35.6 \\ \hline
\textbf{N. Kurdish} & 65.2 & \textbf{74.6} & - & - & - & - & 31.8 & \textbf{39.7} & 8.6 & \textbf{26.7} \\ \hline
\textbf{Nyanja} & 59.7 & \textbf{72.6} & 29.2 & \textbf{37.3} & - & - & 27.6 & \textbf{37.7} & 7.9 & \textbf{29.4} \\ \hline
\textbf{Papiamento} & 75.1 & \textbf{79.6} & - & - & - & - & 52.1 & \textbf{64.4} & 27.8 & \textbf{47.3} \\ \hline
\textbf{Plat. Malagasy} & 64.2 & \textbf{74.1} & 38.9 & \textbf{56.1} & - & - & 31.5 & \textbf{45.0} & 19.3 & \textbf{40.5} \\ \hline
\textbf{Rundi} & 49.8 & \textbf{75.6} & - & - & - & - & 25.3 & \textbf{35.5} & 4.9 & \textbf{19.6} \\ \hline
\textbf{Samoan} & 51.7 & \textbf{75.1} & - & - & - & - & 25.9 & \textbf{40.2} & 8.4 & \textbf{29.9} \\ \hline
\textbf{Scot. Gaelic} & 59.7 & \textbf{64.7} & - & - & - & - & 31.3 & \textbf{44.3} & 10.4 & \textbf{32.3} \\ \hline
\textbf{Shona} & 57.2 & \textbf{79.6} & 35.9 & \textbf{49.1} & 32.8 & \textbf{35.1} & 27.6 & \textbf{40.7} & 8.2 & \textbf{30.4} \\ \hline
\textbf{Sindhi} & 78.6 & \textbf{82.6} & 41.6 & \textbf{55.7} & - & - & 41.0 & \textbf{53.1} & 7.0 & \textbf{33.3} \\ \hline
\textbf{Sinhala} & 58.2 & \textbf{73.6} & 38.7 & \textbf{52.4} & 30.4 & \textbf{32.3} & 29.3 & \textbf{41.7} & 13.2 & \textbf{22.2} \\ \hline
\textbf{S. Azerbaijani} & \textbf{77.1} & 76.1 & - & - & - & - & 32.2 & \textbf{34.7} & \textbf{17.3} & 11.4 \\ \hline
\textbf{S. Pashto} & 69.2 & \textbf{77.1} & 37.0 & \textbf{52.2} & - & - & 39.2 & \textbf{49.4} & 12.4 & \textbf{27.0} \\ \hline
\textbf{S. Sotho} & 57.2 & \textbf{68.7} & 30.8 & \textbf{38.1} & - & - & 26.2 & \textbf{41.5} & 4.3 & \textbf{27.6} \\ \hline
\textbf{Sundanese} & 78.6 & \textbf{81.6} & 44.4 & \textbf{59.5} & - & - & 43.5 & \textbf{53.0} & 23.4 & \textbf{38.7} \\ \hline
\textbf{Swahili} & 73.6 & \textbf{74.6} & 45.2 & \textbf{54.6} & 32.9 & \textbf{35.1} & 41.0 & \textbf{52.8} & 20.1 & \textbf{38.8} \\ \hline
\textbf{Tajik} & 71.6 & \textbf{78.1} & 45.6 & \textbf{59.2} & - & - & 38.7 & \textbf{49.6} & 9.0 & \textbf{31.1} \\ \hline
\textbf{Tatar} & 79.1 & \textbf{83.6} & - & - & - & - & 39.9 & \textbf{48.3} & 10.1 & \textbf{32.0} \\ \hline
\textbf{Telugu} & \textbf{86.6} & 86.1 & \textbf{61.0} & 60.8 & \textbf{38.9} & 38.7 & \textbf{54.9} & 54.2 & 40.2 & \textbf{40.3} \\ \hline
\textbf{Tigrinya} & 23.4 & \textbf{47.8} & 27.6 & \textbf{31.2} & - & - & 16.6 & \textbf{22.8} & 0.4 & \textbf{2.1} \\ \hline
\textbf{Turkmen} & 71.1 & \textbf{83.1} & - & - & - & - & 34.1 & \textbf{45.6} & 13.0 & \textbf{29.0} \\ \hline
\textbf{Uyghur} & 75.1 & \textbf{81.6} & - & - & - & - & 31.9 & \textbf{42.2} & 4.1 & \textbf{24.5} \\ \hline
\textbf{Welsh} & 71.6 & \textbf{76.1} & - & - & - & - & 49.7 & \textbf{62.7} & 21.3 & \textbf{38.0} \\ \hline
\textbf{Xhosa} & 60.7 & \textbf{68.7} & 30.3 & \textbf{41.2} & - & - & 31.0 & \textbf{43.1} & 11.6 & \textbf{26.9} \\ \hline
\textbf{Yoruba} & 62.2 & \textbf{67.2} & 29.8 & \textbf{34.3} & 31.1 & \textbf{33.4} & 32.5 & \textbf{36.9} & 15.7 & \textbf{23.9} \\ \hline
\textbf{Zulu} & 55.7 & \textbf{71.6} & 30.4 & \textbf{41.5} & - & - & 29.5 & \textbf{46.7} & 11.4 & \textbf{28.8} \\ \hline
\end{tabular}%
}
\caption{Full results comparing the base student model (Granite 4 Micro) versus our final trained Kakugo model (denoted \textbf{GenReas+Tran (full)} elsewhere). Values are accuracy for SIB200, Belebele, and GlobalMMLU, and CHRF++ for FLORES.}
\label{tab:full_base_comparison}
\end{table*}

\end{document}